\pdfoutput=1

\documentclass[letterpaper]{article}

\usepackage{caption}
\usepackage{aaai25}
\usepackage{booktabs}
\usepackage{times}
\usepackage{latexsym}

\usepackage[T1]{fontenc}

\usepackage[utf8]{inputenc}

\usepackage{microtype}

\usepackage{inconsolata}

\usepackage{graphicx}

\usepackage{tcolorbox}
\usepackage{xcolor}
\usepackage{natbib}
\usepackage{bibentry}
\usepackage{dsfont}
\usepackage{amsmath}

\DeclareMathOperator{\diag}{diag}

\definecolor{set1-1}{RGB}{228,26,28}  
\definecolor{set1-2}{RGB}{55,126,184} 
\definecolor{set1-3}{RGB}{77,175,74}  
\definecolor{set1-4}{RGB}{152,78,163} 
\definecolor{set1-5}{RGB}{255,127,0}  
\definecolor{set1-6}{RGB}{255,255,51} 
\definecolor{set1-7}{RGB}{166,86,40}  
\definecolor{set1-8}{RGB}{247,129,191}
\definecolor{set1-9}{RGB}{153,153,153}

\definecolor{pastel1-1}{RGB}{251,180,174}
\definecolor{pastel1-2}{RGB}{179,205,227}
\definecolor{pastel1-3}{RGB}{204,235,197}
\definecolor{pastel1-4}{RGB}{222,203,228}
\definecolor{pastel1-5}{RGB}{254,217,166}
\definecolor{pastel1-6}{RGB}{255,255,204}
\definecolor{pastel1-7}{RGB}{229,216,189}
\definecolor{pastel1-8}{RGB}{253,218,236}
\definecolor{pastel1-9}{RGB}{242,242,242}

\title{Towards Symmetric Low-Rank Adapters}

\author{Tales Panoutsos \\
  Department of Computer Science \\
  Federal University of Minas Gerais \\
  Belo Horizonte, Brazil \\
  \texttt{tales.panoutsos@dcc.ufmg.br} \\\And
  Rodrygo L. T. Santos \\
  Department of Computer Science \\
  Federal University of Minas Gerais \\
  Belo Horizonte, Brazil \\
  \texttt{rodrygo@dcc.ufmg.br} \\\And
  Flavio Figueiredo \\
  Department of Computer Science \\
  Federal University of Minas Gerais \\
  Belo Horizonte, Brazil \\
  \texttt{flaviovdf@dcc.ufmg.br} \\
  }

\begin{document}
\maketitle
\begin{abstract}
In this paper, we introduce Symmetric Low-Rank Adapters, an optimized variant of LoRA with even fewer weights. This method utilizes Low-Rank Symmetric Weight Matrices to learn downstream tasks more efficiently. Traditional LoRA accumulates fine-tuning weights with the original pre-trained weights via a Singular Value Decomposition (SVD) like approach, i.e., model weights are fine-tuned via updates of the form $BA$ (where $B \in \mathds{R}^{n\times r}$, $A \in \mathds{R}^{r\times n}$, and $r$ is the rank of the merged weight matrix). In contrast, our approach, named SymLoRA, represents fine-tuning weights as a Spectral Decomposition, i.e., $Q \, diag(\Lambda)\, Q^T$, where $Q \in \mathds{R}^{n\times r}$ and $\Lambda \in \mathds{R}^r$. SymLoRA requires approximately half of the finetuning weights. Here, we show that this approach has negligible losses in downstream efficacy. 
\end{abstract}

\section{Introduction}
Deploying large Natural Language Processing (NLP) services presents significant challenges. End-users of such applications expect fast response times and seamless, conversation-like interactions. Still, the computational demands of evaluating even a pre-trained model can slow down the system, potentially diminishing the user experience. When considering the environmental impact, running these models continuously for such applications can be costly and environmentally taxing. Finally, large models naturally have higher deployment costs.

The cause for such a need are the ever-increasing model sizes in Deep Learning (DL). In fact, there is a belief that larger models may be more suitable for more complex tasks, with NLP tasks being no exception. However, as has been shown by the Pruning Literature~\citep{Blalock2020,Hoefler2021,Wang2022,Wimmer2023}, Deep Neural Networks (NNs) are over-parametrized mainly, with only a fraction of the model weights being needed for accurate predictions. 
Indeed, this is one of the reasons why efficient pruning is sometimes referred to as finding Lottery Tickets, e.g., finding the model's weight that matter and only those weights~\citep{Blalock2020}.

Correlated to the pruning literature, Adapters~\citep{Houlsby2019,Hu2022,Rebuffi2017,Yin2023} are a successful training technique used to fine-tune Large Language Models (LLMs) focusing only on a small fraction of weights. Low-rank adapters (or LoRA) have gained significant traction for language tasks. In a few words, LoRA focuses on saving Low-Rank Singular Value Decomposed (SVDs) updates for model weights. The smaller the SVD rank, the fewer the weights stored. 

Traditional fine-tuning paradigms often involve training part of the parameters or external modules. However, since these methods increase inference latency or reduce the model's usable sequence length, LoRA became a valuable solution for these two problems while presenting several practical vantages. One of these is that different fine-tuned models share the pre-trained weights, making it possible to build small adapter modules, one for each task, containing only the fine-tuning weights and switch easily between them. 

    



In this research paper, we aim at understading if the structure of LoRA updates may be captured by a low-rank Symmetric matrix. In particular, our study focuses on the following hypothesis: {\em May SVD be changed to a Eigen-Decomposition in LoRA?} A natural consequence of this choice, is even more reduction in adapter sizers.


To investigate our hypothesis, we present SymLoRA, a symmetric version of LoRA focused on Eigendecompositions that reduces the number of fine-tuning weights by half.
Reduced model sizes can significantly save storage and operational costs in scenarios where models must be deployed at scale (e.g., large-scale NLP services). 


The proposed method retains all the advantages of LoRA, as it is designed based on the LoRA paradigm. The main contribution of this work is the introduction of symmetric layers. NNs with symmetric layers have been proven to be universal approximators. They are particularly appealing due to their reduced parameter count, which is half that of a conventional matrix. By applying the appropriate decomposition, we can reparametrize the layer to be symmetric and low-rank, achieving even smaller adapters. We evaluate SymLoRA on the Glue benchmark, finding that it is statistically tied to LoRA. Thus, our approach results in little to no performance loss while effectively reducing the number of trainable parameters.

In the next section, we discuss Related Work, followed by our development of SymLoRA, and our Empirical Results. The last section concludes our paper.

\section{Related Work}

This section presents our discussion of related work. It is divided into Low-Rank Adapters, Neural Network Pruning, and Symmetric Neural Networks.


\subsection{Low-Rank Adapters (LoRA)} \label{s:loralow}

Low-Rank Adapters (LoRA) have gained significant attention due to their efficient fine-tuning methods~\citep{Houlsby2019}, which require only a small number of additional parameters. This approach allows the adapter to be merged directly into the original model layer, enabling adaptation to new downstream tasks without introducing any new layers to the NN. Furthermore, the LoRA method provides the flexibility to split the original layer and the adapter, facilitating seamless transitions between different tasks. 

The key insight behind LoRA is that the weight updates of a pre-trained model often exhibit an intrinsic low-rank structure. Consequently, only a low-rank decomposition of the update matrix for the original layer needs to be saved and optimized. Section~\ref{s:method} provides the mathematical details of LoRA. Even though there are several extensions of LoRA, to the best of our knowledge, none have looked into Symmetric NNs as we do.




\subsection{Neural Network Pruning} \label{s:prune}

Neural network pruning is an active research field with a vast amount of literature. We can point out four recent surveys on the subject~\citep{Blalock2020,Hoefler2021, Wang2022,Wimmer2023}. We refer the reader to these surveys for in-depth reviews. 

While SymLoRA is not a running method, it effectively reduces the number of adapter weights that need storage. In this sense, SymLoRA shares the same overarching question of running, which is to find the most miniature model for a given task. Our discussion here will be on the importance of pruning and the main ideas related to SymLoRA.





\subsection{Symmetric Networks} \label{s:sym}

The third related idea is that of exploring symmetric NNs~\citep{Hu2019}. Here, the NN is enforced to have symmetric layers. By definition, if a network is symmetric, it only needs to store upper or lower triangular matrices, thus saving up computing memory. Hu et al.~\citep{Hu2019} showed that symmetric NNs are universal approximators and thus present a promising starting point for pruning. Similarly, some authors propose favoring Spectral Machine Learning~\citep{Giambagli2021,Buffoni2022}. 


\section{SymLoRA} \label{s:method}
In this section, we introduce the proposed method, provide a background overview, and highlight the main differences between LoRA and SymLoRA. Given the well-established foundation of the LoRA method, we begin by elaborating on LoRA and follow on to build SymLoRA.

\subsection{Background}
The LoRA method is applied to pre-trained weights when training a model on a downstream task. In this approach, the pre-trained weights are frozen, and only the LoRA weights are fine-tuned. For a trained and frozen weight matrix $W_0 \in \mathds{R}^{n\times m}$, LoRA optimizes a low-rank matrix to capture the difference between $W_0$ and the fine-tuned matrix for the downstream task. Consequently, during gradient descent, the matrix $W_0$ acts as a constant. The low-rank matrix $\Delta W$ is represented by the product $BA$, where $B \in \mathds{R}^{n\times r}$, $A \in \mathds{R}^{r\times m}$, and $r$ is the rank of the low-rank matrix. The forward pass through this layer is then given by:
$$ h^{(l)} = W_0^{(l)} x + \Delta W^{(l)} x = W^{(l)}_0 x + B^{(l)}A^{(l)}x$$

In our notation, $(l)$ indicates a layer. For simplicity, we drop this index when not required. Note that both matrices, $W_0$ and $\Delta W = BA$, are multiplied by $x$ during the forward pass, but only $\Delta W = BA$ is updated during backpropagation. The matrix $A$ is typically initialized with a random Gaussian distribution, and $B$ is initialized with zeros. This initialization ensures that the model's forward pass is identical to that of the pre-trained model at the beginning of training. Additionally, $\Delta W$ is scaled by $\frac{\alpha}{r}$, where $\alpha$ is a tunable hyperparameter and $r$ is a low rank (user-defined).

It is important to note that since LoRA requires three matrix-vector products in each layer, the latency could be prohibitive. However, in practice, LoRA does not need to be applied to all layers to be effective. Even in the layers where it is applied, the rank $r$ is often much lower than $n$ and $m$, resulting in additional latency comparable to any other fine-tuning prefix-based method.

\subsection{Our Method}
The method proposed here is a variation of the LoRA method that aims to reduce the number of trainable weights further while maintaining the performance and practical benefits of LoRA. Additionally, it expands the exploration of symmetric weight matrices in neural networks. This work's main contribution is to explore the performance of symmetric weight matrices as Low-Rank Adapters.

When LoRA constrains $\Delta W$ to be a low-rank matrix, it represents it via its truncated SVD ($B = U\Sigma$ and $A = V^T$). By doing so, the shape of the matrices constrains the maximum rank of $\Delta W$. In our approach, we use a different decomposition. Specifically, we represent the matrix $\Delta W = Q \diag(\Lambda) Q^T$, where $Q \in \mathds{R}^{n \times r}$ and $\Lambda \in \mathds{R}^{r}$. This not only constrains the rank of the matrix to be at most $r$ but also ensures that it is symmetric. One might argue that $\Delta W$ must be square to be represented by this decomposition. Fortunately, in transformers, most weight matrices are square. LoRA is typically applied only to the square matrices in the attention layers, as demonstrated in the empirical results presented in the LoRA paper. 
\begin{table*}[t!]\centering
\caption{The performance of Lora and SymLora on GLUE benchmark.}\small
\begin{tabular}{lrrrr}\toprule
Datasets & SymLoRA (Val.) & SymLoRA (Test) & LoRA (Val.)\\\midrule
CoLA (Matthew's correlation) &$0.63_{\pm 2.9}$ &$0.60_{\pm -.-}$&$0.63_{\pm -.-}$ \\
QNLI (Accuracy) &$0.92_{\pm 0.7}$ &$0.92_{\pm 0.7}$&$0.93_{\pm 0.3}$ \\
SST-2 (Accuracy) &$0.94_{\pm 1.5}$ &$0.95_{\pm 1.0}$&$0.95_{\pm 0.2}$ \\
MNLI (Accuracy) &$0.86_{\pm 0.5}$ &$0.86_{\pm 0.5}$&$0.87_{\pm 0.3}$ \\
QQP (Accuracy) &$0.90_{\pm 0.3}$ &$0.88_{\pm 0.1}$&$0.90_{\pm 0.1}$ \\
RTE (Accuracy) &$0.85_{\pm 4.1}$ &$0.78_{\pm 1.5}$&$0.86_{\pm 0.7}$ \\
MRPC (Accuracy) &$0.88_{\pm 1.5}$ &$0.88_{\pm 1.5}$&$0.89_{\pm 0.7}$ \\
STS-B (Pearson correlation) &$0.90_{\pm 0.5}$ &$0.88_{\pm -.-}$&$0.91_{\pm -.-}$ \\
\bottomrule
\end{tabular}
\label{t:res}
\end{table*}
\begin{figure*}[t!]
    \centering
    \includegraphics[width=1\linewidth]{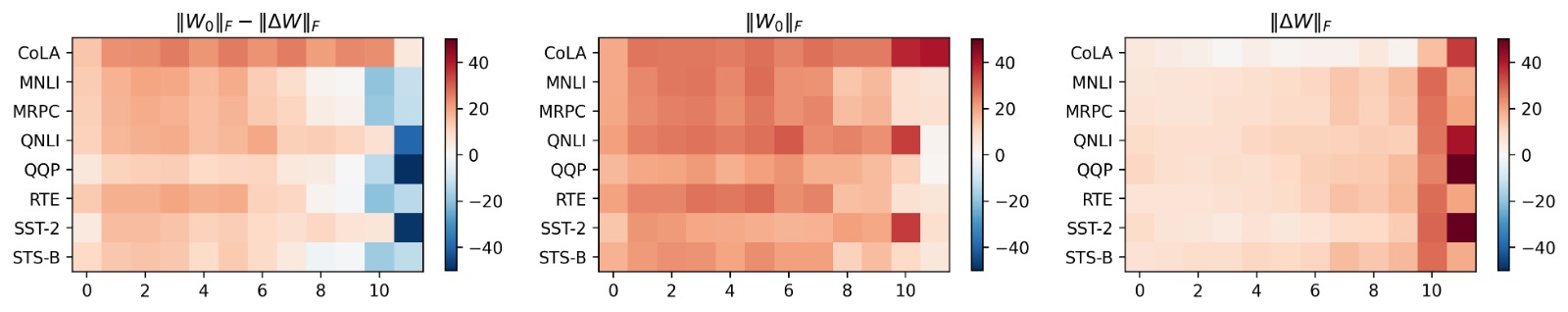}
    \includegraphics[width=\linewidth,trim={0 0 0 5.5cm},clip]{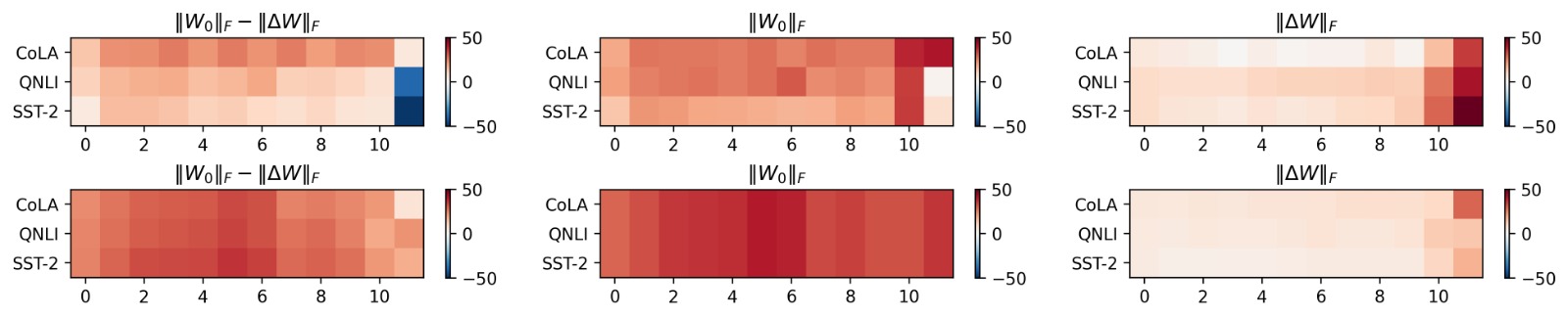}
    \caption{Comparing SymLoRA (top) vs LoRA (bottom). For SymLoRA $\lambda^{(l)} \mid\mid W^{(l)}_0 \mid\mid_f = \lambda^{(l)} \sum_{i,j} \mid W^{(l)}_0(i,j) \mid$. On LoRA we do not have the $\lambda^{(l)}$ argument. All plots are on the same color scale.}
    \label{fig:compare}
\end{figure*}

The forward pass through a Symmetric LoRA layer is defined by:
$$ h^{(l)} = (\lambda^{(l)} W^{(l)}_0) x + (Q^{(l)} \diag(\Lambda^{(l)}) Q^{(l)^T}) x$$
\noindent Again, $(l)$ is the network layer (sometimes dropped from our discussion for simplicity). More importantly, $\lambda^{(l)}$ is also a trainable parameter we add to the method. This parameter serves two purposes. Firstly, it guides whether or not (when the parameter is zero) the downstream tasks need the original weights. Empirically, we find that $\lambda^{(l)}$ is commonly set to zero for the last layers of the NN. Secondly, it enables multiplicative updates (e.g., doubling or adding 10\%) of the original weights. This is not achievable by adding a symmetric matrix to a non-symmetric matrix.

While, LoRA has $2\cdot n\cdot r$ parameters for square matrices, whereas Symmetric LoRA has $(n+1)\cdot r$ parameters. For SymLoRA we also scale the new weights by $\frac{\alpha}{r}$, using the same strategy for choosing $\alpha$ as in LoRA. We will discuss our results next.

\section{Empirical Results}

This section presents our empirical results comparing LoRA with SymLora on downstream tasks from the GLUE benchmark~\citep{Wang2019}. All experiments used RoBERTa~\citep{Liu2019} as the base model. Test set results come from submitting predictions to the GLUE website.

To ensure a fair comparison, we made every effort to use the same hyperparameters and configurations as the LoRA experiments on RoBERTa. We applied SymLora with a rank of $8$ to the value and query matrices of the attention layers, and we set $\alpha$ equal to the rank, just as in LoRA. In addition, we conducted a grid search to determine the best learning rate and batch size for our models. Our results are shown in Table~\ref{t:res}. 

The table shows the average score for each task over validation training sets. The score used in each task is also indicated. Moreover, a 95\% CI. This is the standard $\pm 1.96 \frac{s}{\sqrt{n}}$ confidence interval, where $s$ is the standard deviation. This standard deviation, $s$, is not divulged by the GLUE submission website for test sets. Nevertheless, if the score is a fraction (i.e., accuracy), we estimate it from the formula for the Bernoulli Random Variable, i.e., $s=p(1-p)$, where $p$ is the score. We cannot compute the CI for the test set for scores that are not fractions. Thus, this is omitted. Nevertheless, we do show them for the validation set.



In this table, the values for LoRA were taken from the original LoRA article. To understand this choice, we point out that the main difference between the training of LoRA and SymLoRA was the maximum sequence length, which was $512$ in LoRA and $128$ in our experiments. This change was performed due to GPU memory limitations. To ensure that this change did not impact results, we retrained LoRA on three tasks, namely COLA, QNLI, and SST-2, reaching scores close to the ones reported on the paper, i.e., $63.05\%$ for COLA, $92.71\%$ for QNLI, and $94.95\%$ for SST-2.

Overall, the table presents scores that are close to or tied (when we consider overlapping CIs) with LoRA. The top loss is for RTE. The decrease in accuracy for this task was around $0.8$ in absolute terms and $11\%$ in relative terms (from $0.86\%$ to $0.78$). All other losses are below this one, with most being less than 5\% in relative terms. This points to a slight loss in efficacy when comparing LoRA and SymLoRA.

To better compare SymLoRA and LoRA in Figure~\ref{fig:compare}, we show the difference in Frobenius norms of the original and adapter matrices per task (x-axis) and network layer (y-axis). The top row focuses on SymLoRA, whereas the bottom focuses on LoRA. For the LoRA row, we only show the Frobenius norms of the three tasks we trained locally (COLA, QNLI, and SST-2).

From the top row of the figure, for SymLoRA, we can initially observe a significant difference between the original and adapter norms for most tasks (first column). Notably, this difference is negative for all tasks except COLA. When we examine the norms for the original (middle column) and adapter (last column) weights individually, it is evident that SymLoRA primarily focuses updates on the previous weights (specifically, the last two layers of the last column). Additionally, it is interesting to note that the free parameter $\lambda$ tends to zero out the last layers of the network (seen in the concentration of zeroes in the last layers of the top middle plot).

For LoRA (second row), we can also see this concentration of updates on the last layers (check the previous layers of the last plot on the second row). However, given that LoRA does not have the $\lambda$ parameter, the original weights are not zeroed out as in SymLoRA. As future work, we aim to explore whether this parameter is interesting as an indicator of original layers for deletion.



\section{Conclusion}

In this paper, we present Symmetric Low-Rank Adapters (SymLoRA). SymLoRA represents a significant advancement in parameter efficiency for fine-tuning, surpassing the capabilities of traditional LoRA. This finding indicates that even state-of-the-art fine-tuning methods may not fully optimize parameter utilization. 

Our exploration of layer-wise updates in LoRA and SymLoRA further reveals that the most substantial updates occur predominantly in the final layers. This observation suggests that the new parameters might also be underutilized, warranting further investigations in future research.

Before concluding, we point out that the LoRA literature is already vast~\citep{dettmers2024qlora, valipour2022dylora, hayou2024lora+, Yin2023, Hu2022}. Several authors have extended LoRA to incorporate quantization strategies~\citep{dettmers2024qlora}, dynamic ranks~\citep{valipour2022dylora}, magnitude and direction information~\citep{liu2024dora}, as well as Fourier features~\citep{gao2024parameter}. As future work, we aim at extending these methods to also incorporate symmetric matrices, if possible.

\bibliographystyle{aaai25}
\bibliography{aaai}




\end{document}